\documentclass[11pt,a4paper]{article}
\usepackage[hyperref]{emnlp2018}

\usepackage{times}
\usepackage{latexsym}
\usepackage{graphicx}
\usepackage{url}
\usepackage{tikz}
\usetikzlibrary{patterns, shapes.geometric, positioning, bayesnet}
\usepackage{pgfplots}

\usepackage{amsmath}
\usepackage{subcaption}
\usepackage{verbatim}
\usepackage[none]{hyphenat}
\usepackage{array}
\usepackage{multirow}
\usepackage{makecell}
\usepackage{arydshln}

\usepackage{empheq}

\aclfinalcopy 
\def\aclpaperid{2087} 
\newcolumntype{L}[1]{>{\raggedright\let\newline\\\arraybackslash\hspace{0pt}}m{#1}}
\newcolumntype{C}[1]{>{\centering\let\newline\\\arraybackslash\hspace{0pt}}m{#1}}
\newcolumntype{R}[1]{>{\raggedleft\let\newline\\\arraybackslash\hspace{0pt}}m{#1}}

\title{Challenges of Using Text Classifiers for Causal Inference}

\author{Zach Wood-Doughty$^{*}$, Ilya Shpitser$^\dagger$, Mark Dredze$^{*\dagger}$\\
  Department of Computer Science \\
  $^*$Center for Language and Speech Processing \\
  $^\dagger$Malone Center for Engineering in Healthcare \\
  Johns Hopkins University, Baltimore, MD 21218 \\
  \texttt{\{zach,ilyas,mdredze\}@cs.jhu.edu}}

\begin{document}
\maketitle
\begin{abstract}

Causal understanding is essential for many kinds of decision-making,
but causal inference from observational data
has typically only been applied to structured, low-dimensional datasets.
While text classifiers produce low-dimensional outputs,
their use in causal inference has not previously been studied.
To facilitate causal analyses based on language data,
we consider the role that text classifiers can play in causal inference
through established modeling mechanisms from the causality literature
on missing data and measurement error.
We demonstrate how to conduct causal analyses using text classifiers
on simulated and Yelp data, and discuss the opportunities and challenges
of future work that uses text data in causal inference.

\end{abstract}

\section{Introduction}
Most scientific analyses, in domains from economics to medicine, focus on low-dimensional structured data.
Many such domains also have unstructured text data; advances in natural language processing (NLP)
have led to an increased interest in incorporating language data into scientific analyses.
While language is inherently unstructured and high dimensional, NLP systems can be used to 
process raw text to produce structured variables. 
For example, work on identifying undiagnosed side effects from
electronic health records (EHR) uses text classifiers to produce
clinical variables from the raw text \cite{hazlehurst2009detecting}.

NLP tools may also benefit the study of causal inference, which seeks to identify causal relations from observational data.
Causal analyses traditionally use low-dimensional structured variables, such as clinical markers and binary health outcomes.
Such analyses require assumptions about the data-generating process, which are often simpler with low-dimensional data.
Unlike prediction tasks which are validated by held-out test sets,
causal inference involves modeling counterfactual random variables \cite{neyman23app,rubin76inference}
that represent the outcome of some hypothetical intervention.
To rigorously reason about hypotheticals,
we use causal models to link our counterfactuals to observed data \cite{pearl2009causality}.

NLP provides a natural way to incorporate text data into causal inference models.
We can produce low-dimensional variables using, for example, text classifiers, and then run our causal analysis.
However, this straightforward integration belies several potential issues.
Text classification is not perfect, and errors in a NLP algorithm may bias subsequent analyses.
Causal inference requires understanding how variables influence one another
and how correlations are confounded by common causes.
Classic methods such as stratification provide a means for handling confounding
of categorical or continuous variables, but it is not immediately obvious how such work can be extended to high-dimensional data.

Recent work has approached high-dimensional domains via random forests \cite{wager2017estimation}
and other machine learning methods \cite{chernozhukov2016double}.
But even compared to an analysis that requires hundreds of confounders \cite{belloni2014inference},
NLP models with millions of variables are very high-dimensional.
While physiological symptoms reflect complex biological realities,
many symptoms such as blood pressure are one-dimensional variables.
While doctors can easily quantify the effect of high blood pressure on some outcome,
can we use the ``positivity'' of a restaurant review to estimate a causal effect?
More broadly, is it possible to employ text classification methods in a causal analysis?

We explore methods for the integration of text classifiers into causal inference analyses that consider
confounds introduced by imperfect NLP.
We show what assumptions are necessary for causal analyses using text, and discuss
when those assumptions may or may not be reasonable.
We draw on the causal inference literature to consider two modeling aspects:
missing data and measurement error.
In the missing data formulation, a variable of interest is sometimes unobserved,
and text data gives us a means to model the missingness process.
In the measurement error formulation, we use a text classifier to generate a noisy proxy
of the underlying variable.

We highlight practical considerations of a causal analysis with text data by
conducting analyses with simulated and Yelp data.
We examine the results of both formulations and show how a causal analysis which properly
accounts for possible sources of bias produces better estimates than na\"ive methods which make unjustified assumptions.
We conclude by examining how our approach may enable new research avenues for inferring
causality with text data.

\section{Causal Inference, Briefly}
\begin{figure*}[ht!]
\begin{center}
\begin{subfigure}[b]{0.3\textwidth}
\centering
\begin{tikzpicture}[>=stealth, node distance=1.5cm]
    \tikzstyle{format} = [draw, very thick, circle, minimum size=0.8cm, inner sep=2pt]
    \tikzstyle{unobs} = [draw, very thick, red, circle, minimum size=0.8cm, inner sep=2pt]

    \begin{scope}
        \path[->, very thick]
            node[format] (A) {A}
            node[format, above of=A] (C) {$C$}
            node[format, right of=A] (Y) {$Y$}

            (A) edge[blue] (Y)
            (C) edge[blue] (Y)
            (C) edge[blue] (A)
        ;
    \end{scope}
\end{tikzpicture}
\caption{Simple Confounding}
\label{fig:simple1}
\end{subfigure}
\begin{subfigure}[b]{0.3\textwidth}
\centering
\begin{tikzpicture}[>=stealth, node distance=1.5cm]
    \tikzstyle{format} = [draw, very thick, circle, minimum size=0.8cm, inner sep=2pt]
    \tikzstyle{unobs} = [draw, very thick, red, circle, minimum size=0.8cm, inner sep=2pt]

    \begin{scope}
        \path[->, very thick]
            node[unobs] (A1) {\small $A(1)$}
            node[format, left of=A1] (A) {$A$}
            node[format, above of=A] (RA) {$R_A$}
            node[format, above of=A1] (C) {$C$}
            node[format, right of=A1] (Y) {$Y$}

            (A1) edge[red] (Y)
            (A1) edge[red] (A)
            (C) edge[blue] (A1)
            (C) edge[blue] (Y)
            (C) edge[blue] (RA)
            (Y) edge[blue] (RA)
            (RA) edge[blue] (A)
        ;

    \end{scope}
\end{tikzpicture}
\caption{Missing Data}
\label{fig:missing}
\end{subfigure}
\begin{subfigure}[b]{0.3\textwidth}
\centering
\begin{tikzpicture}[>=stealth, node distance=1.5cm]
    \tikzstyle{format} = [draw, very thick, circle, minimum size=0.8cm, inner sep=2pt]
    \tikzstyle{unobs} = [draw, very thick, red, circle, minimum size=0.8cm, inner sep=2pt]

    \begin{scope}
        \path[->, very thick]
            node[unobs] (A) {$A$}
            node[format, above of=A] (C) {$C$}
            node[format, right of=A] (Y) {$Y$}
            node[format, right of=C] (AA) {$A^*$}

            (A) edge[red] (Y)
            (A) edge[red] (AA)
            (C) edge[blue] (Y)
            (C) edge[blue] (A)
            (C) edge[blue] (AA)
            (Y) edge[blue] (AA)
        ;

    \end{scope}
\end{tikzpicture}
\caption{Measurement Error}
\label{fig:measurement}
\end{subfigure}
\caption{DAGs for causal inference without text data.
Red variables are unobserved.\\
$A$ is a treatment, $Y$ is an outcome, and $C$ is a confounder.}
\end{center}
\label{fig:dags1}
\end{figure*}
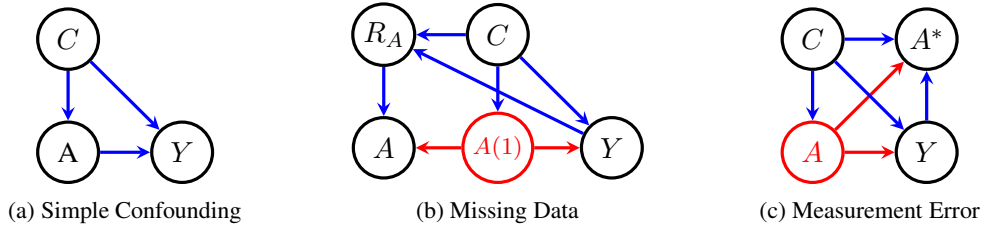

\begin{figure*}[ht!]
\small
\begin{center}
\begin{subfigure}[b]{0.23\textwidth}
\centering

\begin{tabular}{c c c}
$A$ & $C$ & $Y$ \\
\hline
1 & 1 & 0 \\
0 & 1 & 1 \\
0 & 0 & 1 \\
1 & 0 & 1
\end{tabular}

\caption{Simple Confounding}
\label{fig:simple2}
\end{subfigure}
\begin{subfigure}[b]{0.23\textwidth}
\centering

\begin{tabular}{c c c c}
$R_A$ & $A$ & $C$ & $Y$ \\
\hline
1 & 1 & 1 & 0 \\
0 & ? & 1 & 1 \\
1 & 0 & 0 & 1 \\
0 & ? & 0 & 1
\end{tabular}

\caption{Missing Data}
\label{fig:missing2}
\end{subfigure}
\begin{subfigure}[b]{0.23\textwidth}
\centering

\begin{tabular}{c c c}
$A^*$ & $C$ & $Y$ \\
\hline
0 & 1 & 0 \\
0 & 1 & 1 \\
0 & 0 & 1 \\
1 & 0 & 1
\end{tabular}

\caption{Measurement Error}
\label{fig:measurement2}
\end{subfigure}
\begin{subfigure}[b]{0.23\textwidth}
\centering

\begin{tabular}{c c c}
$A^*$ & $A$ \\
\hline
1 & 1  \\
0 & 1  \\
0 & 0  \\
1 & 1 
\end{tabular}
\caption{Mismeasurement}
\label{fig:measurement3}
\end{subfigure}
\end{center}
\caption{Example data rows for causal inference without text data.}
\label{fig:tables}
\end{figure*}

While randomized control trials (RCT) are the gold standard of determining causal effects of treatments on outcomes,
they can be expensive or impossible in many settings. In contrast,
the world is filled with observational data collected without randomization.
While most studies simply report correlations from observational data, the
field of causal inference examines what assumptions and analyses make it possible to identify causal effects.

We formalize a causal statement like ``smoking causes cancer'' as ``if we were to conduct
a RCT and assign smoking as a treatment, we would see
a higher incidence of cancer among those assigned smoking than among the control group.''
In the framework of \newcite{pearl1995causal},
we consider a counterfactual variable of interest: what {\it would have} been
the cancer incidence among smokers if smoking {\it had been} randomized?
Specifically, we consider a causal effect as the counterfactual outcome of a hypothetical intervention
on some treatment variable. If we denote smoking as our treatment variable $A$ and cancer
as our outcome variable $Y$, then
we are interested in the counterfactual distribution, denoted $p(Y(a))$ or $p(Y \mid \textrm{do}(a))$.
We interpret this as ``the distribution over $Y$ had $A$ been set, possibly contrary to fact, to value a.''
For a binary treatment $A$, the causal effect of $A$ on $Y$ is denoted $\tau = E[Y(1)] - E[Y(0)]$;
the average difference between if you had received the treatment and if you had not.
Throughout, we use causal directed acyclic graphs (DAG),
which assumes that an intervention on $A$ is well-defined and results in
a counterfactual variable $Y(a)$ \cite{pearl1995causal,dawid2010beware}.


Figure \ref{fig:simple1} shows an example of simple confounding.
This is the simplest DAG in which counterfactual distribution $p(Y(a))$ is not simply $p(Y \mid A)$,
as $C$ influences both the treatment $A$ and the outcome $Y$.
To recover the counterfactual distribution $p(Y(a))$ that would follow an intervention upon $A$,
we must ``adjust'' for $C$, applying the so-called ``back-door criterion'' \cite{pearl1995causal}.
We can then derive the counterfactual distribution $p(Y(a))$
and desired causal effect, $\tau_S$ as a function of the observed data,
(Fig. \ref{fig:equations} Eq. \ref{eq:simple}.) This derivation is shown in Appendix \ref{app:simple}.

Note that $p(Y(a))$ and $\tau_S$ require data on $C$, and if $C$ is not in fact observed,
it is impossible to recover the causal effect.
Formally, we say that $p(Y(a))$ is {\it not identified} in the model,
meaning there is no function $f$ such that $p(Y(a))$$=$$f(p(A, Y))$.
Identifiability is a primary concern of causal inference \cite{shpitser2008complete}.

Throughout, we assume for simplicity that $A$, $C$, and $Y$ are binary variables.
While text classifiers can convert high-dimensional data into binary variables for such analyses,
we need to make further assumptions about how classification errors affect causal inferences.
We cannot assume that the output of a text classifier can be treated as if it were ground truth.
To conceptualize the ways in which a text classifier may be biased, we will consider them as a way to
recover from missing data or measurement error.

\section{Causal Models}

Real-world observational data is messy and often imperfectly collected.
Work in causal inference has studied how analyses can be made robust to missing data
or data recorded with systematic measurement errors.

\subsection{Missing Data} \label{subsec:missing}

Our dataset has ``missing data'' if it contains individuals (instances) for which some variables are unobserved,
even though these variables are typically available. This may occur if some survey respondents
choose not to answer certain questions,
or if certain variables are difficult to collect and thus infrequently recorded. 
Missing data is closely related to causal inference -- both are interested in
hypothetical distributions that we cannot directly observe \cite{robins2000sensitivity,shpitser2015missing}.

Consider a causal model where $A$ is sometimes missing (Figure \ref{fig:missing}).
The variable $R_A$ is a binary indicator for whether $A$ is observed ($R_A = 1$) or missing.
The variable $A(R_A = 1$), written as $A(1)$, represents the counterfactual value of $A$ were it never missing.
Finally, $A$ is the observed proxy for $A(1)$: it has the same value as $A(1)$ if $R_A = 1$,
and the special value ``?'' if $R_A = 0$.

Solving missingness can seen as intervening to set $R_{A}$ to 1.
Given $p(A, R_{A}, C, Y)$, we want to recover $p(A(1), C, Y)$.
We may need to make a ``Missing at Random'' (MAR) assumption,
which says that the missingness process is independent of the true missing
values, conditional on observed values.
Figure \ref{fig:missing} reflects the MAR assumption;
$R_A$ is independent of $A(1)$ given fully-observed $C$ and $Y$.
If an edge existed from $A(1)$ to $R_{A}$,
we have ``Missing Not at Random'' (MNAR)
and would not be identified except in special cases \cite{shpitser2015missing}.

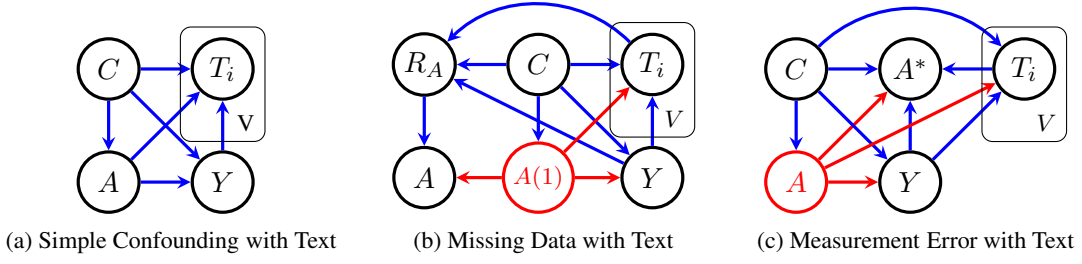
\begin{figure*}[t!]
\begin{center}
\begin{subfigure}[b]{0.3\textwidth}
\centering
\begin{tikzpicture}[>=stealth, node distance=1.5cm]
    \tikzstyle{format} = [draw, very thick, circle, minimum size=0.8cm, inner sep=2pt]
    \tikzstyle{unobs} = [draw, very thick, red, circle, minimum size=0.8cm, inner sep=2pt]

    \begin{scope}
        \path[->, very thick]
            node[format] (A) {$A$}
            node[format, above of=A] (C) {$C$}
            node[format, right of=A] (Y) {$Y$}
            node[format, right of=C] (T) {$T_i$}

            (A) edge[blue] (Y)
            (C) edge[blue] (Y)
            (C) edge[blue] (A)
            (C) edge[blue] (T)
            (A) edge[blue] (T)
            (Y) edge[blue] (T)
        ;
        \plate {plate1} {(T)} {V};
    \end{scope}
\end{tikzpicture}
\caption{Simple Confounding with Text}
\label{fig:simple-text}
\end{subfigure}
\begin{subfigure}[b]{0.3\textwidth}
\centering
\begin{tikzpicture}[>=stealth, node distance=1.5cm]
    \tikzstyle{format} = [draw, very thick, circle, minimum size=0.8cm, inner sep=2pt]
    \tikzstyle{unobs} = [draw, very thick, red, circle, minimum size=0.8cm, inner sep=2pt]

    \begin{scope}
        \path[->, very thick]
            node[unobs] (A1) {\small $A(1)$}
            node[format, left of=A1] (A) {$A$}
            node[format, above of=A] (RA) {$R_A$}
            node[format, above of=A1] (C) {$C$}
            node[format, right of=A1] (Y) {$Y$}
            node[format, right of=C] (T) {$T_i$}

            (A1) edge[red] (Y)
            (A1) edge[red] (A)
            (C) edge[blue] (A1)
            (C) edge[blue] (Y)
            (C) edge[blue] (RA)
            (Y) edge[blue] (RA)
            (RA) edge[blue] (A)

            (C) edge[blue] (T)
            (A1) edge[red] (T)
            (Y) edge[blue] (T)
            (T) edge[blue, bend right=45] (RA)
        ;
        \plate {plate1} {(T)} {$V$};

    \end{scope}
\end{tikzpicture}
\caption{Missing Data with Text}
\label{fig:missing-text}
\end{subfigure}
\begin{subfigure}[b]{0.3\textwidth}
\centering
\begin{tikzpicture}[>=stealth, node distance=1.5cm]
    \tikzstyle{format} = [draw, very thick, circle, minimum size=0.8cm, inner sep=2pt]
    \tikzstyle{unobs} = [draw, very thick, red, circle, minimum size=0.8cm, inner sep=2pt]

    \begin{scope}
        \path[->, very thick]
            node[unobs] (A) {$A$}
            node[format, above of=A] (C) {$C$}
            node[format, right of=A] (Y) {$Y$}
            node[format, right of=C] (AA) {$A^*$}
            node[format, right of=AA] (T) {$T_i$}

            (A) edge[red] (Y)
            (A) edge[red] (AA)
            (C) edge[blue] (Y)
            (C) edge[blue] (A)
            (C) edge[blue] (AA)
            (Y) edge[blue] (AA)

            (C) edge[blue, bend left=45] (T)
            (A) edge[red] (T)
            (Y) edge[blue] (T)
            (T) edge[blue] (AA)
        ;
        \plate {plate1} {(T)} {$V$};

    \end{scope}
\end{tikzpicture}
\caption{Measurement Error with Text}
\label{fig:measurement-text}
\end{subfigure}
\end{center}
\caption{DAGs for causal inference with text data.
In the Yelp experiments we discuss, $T_i$ influences $Y$ and not the other way around.}
\label{fig:dags2}
\end{figure*}

\begin{figure*}
{\small
\begin{align}
\tau_S = \sum_C \left( p(Y=1 \mid A=1, C) - p(Y=1 \mid A=0, C) \right) p(C)
\label{eq:simple}
\end{align}
}

\vspace{-3em}

{\small
\begin{align}
\tau_\text{MD} = \sum_C \bigg( &\dfrac{p(A=1 \mid {\bf T}, C, Y=1, R_A = 1)}
{\sum_y' p(A = 1 \mid {\bf T}, C, y', R_A = 1)p(Y=y' \mid C)} \nonumber \\
&- \dfrac{p(A = 0 \mid {\bf T}, C, Y=1, R_A = 1)}
{\sum_y' p(A = 0 \mid {\bf T}, C, Y=y', R_A = 1)p(Y=y' \mid C)} \bigg) p(Y=1, C)
\label{eq:missing}
\end{align}
}

\vspace{-2em}

{ \small
\begin{align}
\tau_\text{ME} = \sum_C \left( \dfrac{\dfrac{-\delta_{c, y=1}q_{c,y=1}(0) + (1 - \delta_{c,y=1})q_{c,y=1}(1)}{(1 - \epsilon_{c, y=1} - \delta_{c, y=1})}}{\sum_{y'}\dfrac{-\delta_{c, y'}q_{c,y'}(0) + (1 - \delta_{c,y'})q_{c,y'}(1)}{(1 - \epsilon_{c, y'} - \delta_{c, y'})}}
-  \dfrac{\dfrac{(1-\epsilon_{c, y=1})q_{c,y=1}(0) - \epsilon_{c,y=1} q_{c,y=1}(1)}{(1 - \epsilon_{c, y=1} - \delta_{c, y=1})}}{\sum_{y'}\dfrac{(1-\epsilon_{c, y'})q_{c,y'}(0) - \epsilon_{c,y'} q_{c,y'}(1)}{(1 - \epsilon_{c, y'} - \delta_{c, y'})}} \right) p(C)
\label{eq:measurement}
\end{align}
\begin{center}
Define $\epsilon_{c, y} = p(A=0 \mid A^* = 1, C=c, Y=y)$, 
$\delta{c, y} = p(A=1 \mid A^* = 0, C=c, Y=y)$,\\
$q_{c,y}(0)  = p(C=c, Y=y, A^* = 0)$, and
$q_{c,y}(1)  = p(C=c, Y=y, A^* = 1)$.
\end{center}
}
\caption{Functionals for the Causal Effects for Simple Confounding ($\tau_\text{SC}$), Missing Data ($\tau_\text{MD}$) and Measurement Error ($\tau_\text{ME}$).
Derivations are in Appendices \ref{app:simple}, \ref{app:missing}, and \ref{app:measurement}.}
\label{fig:equations}
\end{figure*}

\subsection{Measurement Error} \label{subsec:measurement}

Sometimes a necessary variable is never observed,
but is instead proxied by a variable which differs from the truth by some error.
Consider the example of body mass index (BMI) as a proxy for obesity in a clinical study.
Obesity is a known risk factor for many health outcomes, but has a complex clinical definition and is nontrivial to measure.
BMI is a simple deterministic function of height and weight. To conduct a causal analysis of obesity
on cancer when only BMI and cancer are measured, we can proceed as if we had
measured obesity and then correct our analysis for the known error that comes from using BMI
as a proxy for obesity \cite{hernan2009invited,michels1998does}.

To generalize this concept, we can replace obesity with our ground truth variable $A$ and replace BMI
with a noisy proxy $A^*$. Figure \ref{fig:measurement} gives the DAG for this model.
Unlike missing data problems, there is no hypothetical intervention which recovers
the true data distribution $p(A, C, Y)$. Instead, we manipulate the observed
distribution $p(A^*, C, Y)$ with the known relationship $p(A^*, A)$ to recover the desired
$p(A, C, Y)$.

Unlike missing data, measurement error conceptualization can be used even when
we never observe $A$ (e.g. the table in Figure \ref{fig:measurement2})
as long as we have knowledge about the error mechanism $p(A^*, A)$.
Using this knowledge, we can correct for the error
using `matrix adjustment' \cite{pearl2010measurement}.
In practice we might learn $p(A^*, A)$ from data such as that found in Figure \ref{fig:measurement3}.
Recent work has also considered how multiple independent
proxies of $A$ could allow identification without
any data on $p(A^*, A)$ \cite{kuroki2014measurement}.

\section{Causal Models for Text Data}

We can use conceptualizations for missing data and measurement error to support
causal analyses with text data.
The choice of model depends on the assumptions we make about the data-generation process.

We add new variables to our models (Figure \ref{fig:simple1}) to represent text, which
produces the data-generating distribution shown in Figure \ref{fig:simple-text}.
This model assumes that the underlying $A$, $C$, and $Y$ variables are generated before the text variables;
we use text to recover the true relationship between $A$ and $Y$.

We represent text as an arbitrary set of $V$ variables, which are independent of one another
given the non-text variables. In our implemented analyses we will represent
text as a bag-of-words, wherein each $T_{i}$ is simply the binary indicator of the presence of the $i$-th word
in our vocabulary of $V$ words, and ${\bf T} = \cup_i T_i$.
The restriction to simple text models allows us to explore connections
to causal inference applications, though future work could relax assumptions of the text models 
to be inclusive of more sophisticated text models (e.g. neural sequence models \cite{lai2015recurrent,zhang2015character}),
or consider causal relationships between two text variables.

To motivate our explanations, consider the task of predicting an individuals' smoking status from free-text
hospital discharge notes \cite{uzuner2008identifying,wicentowski2008using}. Some hospitals do not explicitly
record patient smoking status as structured data, making it difficult to use such data in a study on the outcomes
of smoking. We will suppose that we are given a dataset with patient data on lung cancer outcome ($Y$) and
age ($C$), that our data on smoking status ($A$) is affected by either missing data or measurement error,
but that we have text data (${\bf T}$) from discharge records that will allow us to infer smoking status with
reasonable accuracy.

\subsection{Missing Data} \label{subsec:text-as-missing}

To show how we might use text data to recover from missing data,
we introduce missingness for $A$ from Figure \ref{fig:simple-text}
to get the model in Figure \ref{fig:missing-text}.
The missing arrow from $A(1)$ to $R_A$ encodes the MAR assumption,
which is sufficient to make it possible to identify the full data distribution from the observed data.

Suppose our motivation is to estimate the causal effect of smoking status ($A$) on lung cancer ($Y$) adjusting for age ($C$).
Imagine that missing data arises because hospitals sometimes -- but not always -- delete explicit data on smoking status from patient records.
If we have access to patients' discharge notes (${\bf T}$) and know whether a given patient had smoking status recorded ($R_A$), then
the DAG in Figure \ref{fig:missing-text} may be a reasonable model for our setting.
Note that we must again assume that $A$ does not directly affect $R_A$.

The causal effect of $A$ on $Y$ in Figure \ref{fig:missing-text} is identified
as $\tau_{MD}$, given in Eq.~\ref{eq:missing} in Figure \ref{fig:equations}.
The derivation is given in Appendix \ref{app:missing}.

\subsection{Measurement Error} \label{subsec:text-as-measurement}

We model text data with measurement error by introducing
a proxy $A^*$ to the model in Figure \ref{fig:measurement-text}.
We assume that the proxied value of $A^*$ can depend upon all other variables,
and that we will be able to estimate $p(A^*, A)$ given an external dataset, e.g. text classifier
accuracy on held-out data.

Suppose we again want to estimate the causal effect from \S \ref{subsec:text-as-missing}, but this time
none of our hospital records contain explicit data on smoking status.
However, imagine that we have a separate training dataset of medical discharge records
annotated by expert pulmonologists for patients' smoking status.
We could then train a classifier to predict smoking status using discharge record
text\footnote{This is the precise setting of \newcite{uzuner2008identifying}.}.

Working from the derivation for matrix adjustment in binary models given by
\newcite{pearl2010measurement}, we identify
the causal effect of $A$ on $Y$ (Figure \ref{fig:measurement-text}) as $\tau_\text{ME}$
(Eq \ref{eq:measurement} in Figure \ref{fig:equations}.)
The derivation is in Appendix \ref{app:measurement}.

\section{Experiments} \label{sec:experiments}

We now empirically evaluate the effectiveness of our two conceptualizations (missing data and measurement error)
for including text data in causal analyses.
We induce missingness or mismeasurement of the treatment
variable and use text data to recover the true causal relationship of
that treatment on the outcome.
We begin with a simulation study with synthetic text data,
and then conduct an analysis using reviews from \texttt{yelp.com}.

\subsection{Synthetic Data}
We select synthetic data so that we can control the entire data-generation process.
For each data row, we first sample data on three binary variables ($A$, $C$, $Y$) and then 
sample $V$ different binary variables $T_i$ representing a $V$-vocabulary bag-of-words. 
A graphical model for this distribution appears in Figure \ref{fig:simple-text}.
We augment this distribution to introduce either missing data (Figure \ref{fig:missing-text})
or measurement error (Figure \ref{fig:measurement-text}.)
For measurement error, we sample two datasets.
A small training set which gives data on $p(A, C, Y, {\bf T})$
and a large test set which gives data on $p(C, Y, {\bf T})$.

The full data generating process appears in Appendix \ref{app:synthetic},
and the implementation (along with all our code) is provided
online\footnote{\texttt{github.com/zachwooddoughty/emnlp2018-causal}}.

\subsection{Yelp Data}

We utilize the 2015 Yelp Dataset Challenge\footnote{\texttt{yelp.com/dataset/challenge}}
which provides 4.7M reviews of local businesses.
Each review contains a one- to five-star rating, up to 5,000 characters of text.
Yelp users can flag reviews as ``Useful'' as a mark of quality.

We extract treatment, outcome, and confounder variables from the structured data.
The treatment is a binarized user rating that takes value 1
if the review has four or five stars and value 0 if the review has one or two stars.
Three-star reviews are discarded from our analysis.
The outcome is whether the review received at least one ``Useful'' flag.
The confounder is whether the review's author has received at least two ``Useful'' flags
across all reviews, according to their user object.
In our data, 74.2\% of reviews were positive, 42.6\% of reviews were flagged
as ``Useful,'' and 56.7\% users had received at least two such flags.
We preprocess the text of each review by lowercasing, stemming, and removing stopwords,
before converting to a bag-of-words representation with the 4,334 word vocabulary
of all words which appeared at least 1000 times in a sample of 1M reviews.

Based on this $p(A, C, Y, {\bf T})$ distribution, we 
assume the data-generating process that matches Figure \ref{fig:simple-text} and introduce
missingness and mismeasurement as before, giving us data-generating
processes matching Figures \ref{fig:missing-text} and \ref{fig:measurement-text}.

Our intention is not to argue about a true real-world causal effect of Yelp reviews on
peer behavior: we do not believe that our confounder is the only common cause
of the author's rating and the platform's response.
We leave for future work a case study that jointly addresses questions of identifiability
and estimation of a real-world causal effect.
In this work, our experiments focus on a simpler task:
can a correctly-specified model that uses text data 
effectively estimate a causal effect in the presence of missing data
or measurement error.

\subsection{Models}
We now introduce several baseline methods which, unlike our correctly specified models $\tau_{MD}$ and $\tau_{ME}$,
are not consistent estimators of our desired causal effect. 
We would expect that the theoretical bias in these estimators would result in poor performance
in our experiments.

\subsubsection{Baseline: Na\"ive Model}

In both the missing data and measurement error settings, our models use some rows that are full observed.
In missing data, these are rows where $R_A = 1$; in measurement error, the training set is sampled
from the true distribution. The simplest approach to handling imperfect data is to throw away all rows without full data,
and calculate Eq \ref{eq:simple} from that data. In Figure \ref{fig:results}, these are labeled as \texttt{*.naive}.

\subsubsection{Baseline: Textless Model}

In Figure \ref{fig:missing-text}, if we do not condition on $T_i$ to d-separate $A(1)$ from its missingness indicator,
that influence may bias our estimate.
While we know that ignoring text may introduce asymptotic bias into our estimates of the causal effect, we empirically evaluate
how much bias is produced by this ``Textless'' model compared to a correct model. This is labeled as \texttt{*.no\_text}
in Figure \ref{fig:results} (a).

In principle, we could conduct a measurement error analysis using a model that does not
include text. In practice, we found we could not impute $A^*$ from $C$ and $Y$ alone.
The non-textual classifier had such high error that the adjustment matrix was singular and we could
not compute the effect. Thus, we have no such baseline in our measurement error results.

\subsubsection{Baseline: \texttt{no\_y} and \texttt{unadjusted} Models}

In Figure \ref{fig:missing-text}, we must also condition on $C$ and $Y$ to d-separate $A(1)$ from its missingness indicator.
In our misspecified model for missing data, we do not condition on $Y$, leaving open a path for $A(1)$ to influence its missingness.
In Figure \ref{fig:results} (a), this model is labeled as \texttt{*.no\_y}.

When correcting for measurement error, a crucial piece of the estimation is the matrix adjustment using the known
error between the proxy and the truth. A straightforward misspecified model for measurement error is
to impute a proxy for each row in our dataset
and then calculate the causal effect assuming no error between the proxy and truth.
This approach, while simplistic, can be thought of as using a text classifier as a proxy
without regard for the text classifier's biases.
In Figure \ref{fig:results} (b), this approach is labeled as \texttt{*.unadjusted}.

\subsubsection{Correct Models}

Finally, we consider the estimation approaches presented in \S \ref{subsec:text-as-missing} and \S \ref{subsec:text-as-measurement}.
For the missing data causal effect ($\tau_\text{MD}$ from Eq \ref{eq:missing})
we use a multiple imputation estimator which calculates the average effect
across 20 samples from $p(A | {\bf T}, C, Y)$ for each row where $R_A = 0$.
For the measurement error causal effect ($\tau_\text{ME}$ from Eq \ref{eq:measurement}),
we use the training set of $p(A, C, Y, {\bf T})$ data to estimate $\epsilon_{c, y}$ and $\delta{c, y}$
and the larger set of $p(C, Y, {\bf T})$ data to estimate $q_{c,y}$ and $p(C)$.

These models are displayed in Figure \ref{fig:results} (a) as \texttt{*.full} and in Figure \ref{fig:results} (b) \texttt{*.adjusted}.

\subsection{Evaluation}

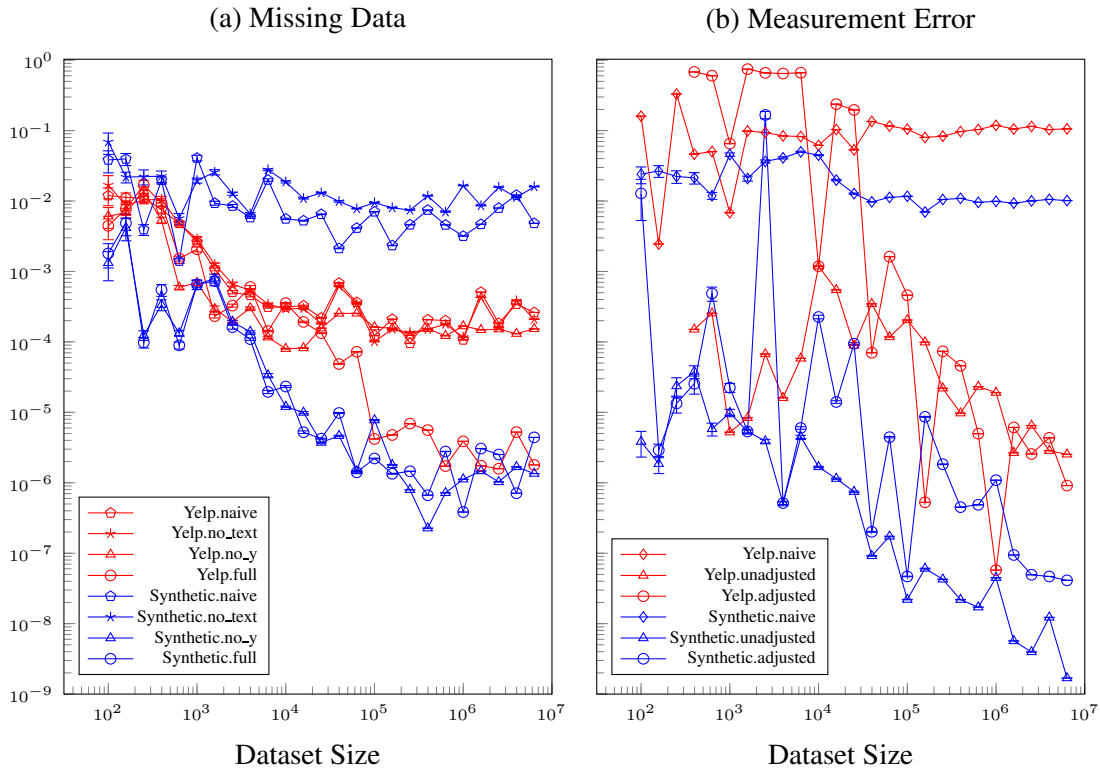
\begin{figure*}[t]
\begin{center}
\centering
\begin{tikzpicture}
\begin{loglogaxis}[
title={(a) Missing Data},
xlabel={Dataset Size},
height=10cm,
width=8cm,
ymin=.000000001,
ymax=1.03,
xmax=1e7,
ytick pos=left,
xtick pos=left,
ticklabel style={font=\tiny},
legend entries={
                Yelp.naive, Yelp.no\_text, Yelp.no\_y, Yelp.full,
                Synthetic.naive, Synthetic.no\_text, Synthetic.no\_y, Synthetic.full,
                },
legend style={ nodes={scale=0.6, transform shape}, cells={anchor=east}, legend pos=south west,},
]
\addplot [color=red, mark=pentagon,]
 plot [error bars/.cd, y dir = both, y explicit]
 table[x =n, y =err, y error =se]{arxiv_dats/md.yelp.1000.naive.dat};
\addplot [color=red, mark=star,]
 plot [error bars/.cd, y dir = both, y explicit]
 table[x =n, y =err, y error =se]{arxiv_dats/md.yelp.1000.textless.dat};
\addplot [color=red, mark=triangle,]
 plot [error bars/.cd, y dir = both, y explicit]
 table[x =n, y =err, y error =se]{arxiv_dats/md.yelp.1000.bad_mi.dat};
\addplot [color=red, mark=o,]
 plot [error bars/.cd, y dir = both, y explicit]
 table[x =n, y =err, y error =se]{arxiv_dats/md.yelp.1000.mi.dat};
\addplot [color=blue, mark=pentagon,]
 plot [error bars/.cd, y dir = both, y explicit]
 table[x =n, y =err, y error =se]{arxiv_dats/md.synthetic.1000.naive.dat};
\addplot [color=blue, mark=star,]
 plot [error bars/.cd, y dir = both, y explicit]
 table[x =n, y =err, y error =se]{arxiv_dats/md.synthetic.1000.textless.dat};
\addplot [color=blue, mark=triangle,]
 plot [error bars/.cd, y dir = both, y explicit]
 table[x =n, y =err, y error =se]{arxiv_dats/md.synthetic.1000.bad_mi.dat};
\addplot [color=blue, mark=o,]
 plot [error bars/.cd, y dir = both, y explicit]
 table[x =n, y =err, y error =se]{arxiv_dats/md.synthetic.1000.mi.dat};
\end{loglogaxis}
\end{tikzpicture}
\hspace{-0.5em}
\begin{tikzpicture}
\begin{loglogaxis}[
title={(b) Measurement Error},
xlabel={Dataset Size},
height=10cm,
width=8cm,
ymin=.000000001,
ymax=1.03,
xmax=1e7,
ytick pos=left,
yticklabels={,,},
xtick pos=left,
ticklabel style={font=\tiny},
legend entries={Yelp.naive, Yelp.unadjusted, Yelp.adjusted,
                Synthetic.naive, Synthetic.unadjusted, Synthetic.adjusted},
legend style={ nodes={scale=0.6, transform shape}, cells={anchor=east}, legend pos=south west,},
]
\addplot [color=red, mark=diamond,]
 plot [error bars/.cd, y dir = both, y explicit]
 table[x =n, y =err, y error =se]{arxiv_dats/me.yelp.1000.naive.dat};
\addplot [color=red, mark=triangle,]
 plot [error bars/.cd, y dir = both, y explicit]
 table[x =n, y =err, y error =se]{arxiv_dats/me.yelp.1000.misspecified.dat};
\addplot [color=red, mark=o,]
 plot [error bars/.cd, y dir = both, y explicit]
 table[x =n, y =err, y error =se]{arxiv_dats/me.yelp.1000.correct.dat};
\addplot [color=blue, mark=diamond]
 plot [error bars/.cd, y dir = both, y explicit]
 table[x =n, y =err, y error =se]{arxiv_dats/me.synthetic.1000.naive.dat};
\addplot [color=blue, mark=triangle,]
 plot [error bars/.cd, y dir = both, y explicit]
 table[x =n, y =err, y error =se]{arxiv_dats/me.synthetic.1000.misspecified.dat};
\addplot [color=blue, mark=o,]
 plot [error bars/.cd, y dir = both, y explicit]
 table[x =n, y =err, y error =se]{arxiv_dats/me.synthetic.1000.correct.dat};
\end{loglogaxis}
\end{tikzpicture}
\end{center}
\caption{Experimental results. Squared distance (y-axis, lower is better) of the estimated causal effect from
$\tau_\text{SC}$ calculated from the full data with no missing data or measurement error.
Error bars (negligible for larger datasets) are 1.96 times standard error across 10 experiments.
Additional experiments with a larger vocabulary are shown in Appendix \ref{app:experiments}. \label{fig:results}}
\end{figure*}

Each model takes in a data sample with missingness or mismeasurement,
and outputs an estimate of the causal effect of A on Y in the underlying data.
Rather than comparing models' estimates against a population-level estimate,
we compare against an estimate of the effect computed on the same data sample, but without
any missing data or measurement error. This `perfect data estimator' may still make errors
given the finite data sample. We compare against this estimator to avoid a small-sample
case where an estimator gets lucky.
In Figure \ref{fig:results}, we plot data sample size against
the squared distance of each model's estimate
from a perfect data estimator's estimate, averaged over ten runs.
Figure \ref{fig:results2} in Appendix \ref{app:experiments} contains
a second set of experiments using a larger vocabulary.

\section{Results}

Given that our correctly-specified models are proven to be asymptotically consistent,
we would expect them to outperform misspecified models.
However, for any given dataset, asymptotic consistency provides no guarantees.

\subsection{Missing Data}

The missing data (MD) experiments suggest that the correct \texttt{full} model does perform best.
The \texttt{no\_y} model performs approximately as well as the correct model on the synthetic data,
but not on the Yelp data. 
The difference between the \texttt{no\_y} and \texttt{full} missing data models is simply a function
of the effect of $Y$ on $R_A$. We could tweak our synthetic data distribution to increase the influence of $Y$
to make the \texttt{no\_y} model perform worse.

When we initially considered other data-generating distributions for missing data,
we found that when we reduced the influence of
the text variables on $R_A$, the \texttt{no\_text} and \texttt{naive} models approached the performance of
the correctly-specified model. While intuitive, this reinforces that the underlying distribution
matters a great deal in how modeling choices may introduce biases if incorrectly specified.

\subsection{Measurement error}

The measurement error results tell a more interesting story. We see enormous fluctuations of the
\texttt{adjusted} model, and in the synthetic data, the \texttt{unadjusted} model appears to be
quite superior.

In the synthetic dataset, this is likely because our text classifier had near-perfect accuracy,
and so simple approach of assuming its predictions were ground-truth introduced less bias.
A broader issue with the \texttt{adjusted} model is that the matrix adjustment approach requires
dividing by (potentially very small) probabilities, this sometimes resulted in huge over-corrections.
In addition, since those probabilities are
estimated from a relatively small training dataset,
small changes to the error-estimate can propagate to huge changes in the final casual estimate.

This instability of the matrix adjustment approach may be a bigger problem for text and other
high-dimensional data: unlike in our earlier example of BMI and obesity, there are likely no simple relationships
between text and clinical variables. However, instead of using matrix adjustment as a way to recover
the true effect, we may instead use it to bound the error our proxy may introduce.
As mentioned by \newcite{pearl2010measurement}, when $p(A \mid A^*)$ is not known exactly,
we can use a Bayesian analysis to bound estimates of a causal effect. In a downstream task, this would
let us explore the stability of our \texttt{adjusted} results.

\section{Related Work}
A few recent papers have considered the possibilities for combining text data with approaches
from the causal inference literature.
\newcite{landeiro2016robust} and \newcite{landeiro2017controlling} explored text classification
when the relationship between text data and class labels are confounded.
Other work has used propensity scores as a way to extract features from text data \cite{paul2017feature}
or to match social media users based on what words they write \cite{de2016discovering}.
The only work we know of which seeks to estimate causal effects using text data focuses on
effects {\it of} text or effects {\it on} text \cite{egami2018make,roberts2018adjusting}. In our work,
our causal effects do not include text variables: we use text variables to recover an underlying
distribution and then estimate a causal effect within that distribution.

There is a conceptually related line of work in the NLP community on inferring 
causal relationships expressed in text \cite{girju2003automatic,kaplan1991knowledge}.
However, our work is fundamentally different. Rather than identify casual relations expressed via language,
we are using text data in a causal model to identify the strength of an underlying causal effect.

\section{Future Directions}
While this paper addresses some initial issues arising from
using text classifiers in causal analyses, many challenges remain.
We highlight some of these issues as directions for future research.

We provided several proof-of-concept models for estimating effects,
but our approach is flexible to more sophisticated models.
For example, a semi-parametric estimator would make no assumptions
about the text data distribution by wrapping the text
classifier into an infinite-dimensional
nuisance model \cite{tsiatis2007semiparametric}.
This would enable estimators robust to partial model
misspecification \cite{bang2005doubly}.

Choices in the design of statistical models of text consider issues
like accuracy and tractability. Yet if these models are to be used
in a causal framework, we need to understand how modeling assumptions 
introduce biases and other issues that can interfere with a downstream
causal analysis. To take an example from the medical domain,
we know that doctors write clinical notes throughout the healthcare process,
but it is not obvious how to model this data-generating process.
We could assume that the doctor's notes passively record a patient's progression,
but in reality it may be that the content of the notes themselves
actively change the patient's care; causality could work in either direction.

New lines of work in causality may be especially helpful for NLP. 
In this work, we used simple logistic regression on a bag-of-words representation
of text; using state-of-the-art text models will likely require more causal
assumptions.
\newcite{nabi2017semi} develops causality-preserving dimensionality reduction, which
could help develop text representations that preserve causality.

Finally, we are interested in case studies on incorporating text classifiers into real-world causal analyses.
Many health studies have used text classifiers to extract clinical variables from EHR data \cite{meystre2008extracting}.
These works could be extended to study causal effects involving those extracted variables,
but such extensions would require an understanding of the underlying assumptions. 
In any given study, the necessity and appropriateness of assumptions will hinge on domain expertise.
The conceptualizations outlined in this paper, while far from solving all issues of causality and text,
will help those using text classifiers to more easily consider research questions of cause and effect.

\section*{Acknowledgments}

This work was in part supported by the National Institute of General
Medical Sciences under grant number 5R01GM114771 and by
the National Institute of Allergy and Infectious Diseases under
grant number R01 AI127271-01A1.
We thank the anonymous reviewers for their helpful comments.

\bibliography{arxiv_paper}
\bibliographystyle{acl_natbib}

\clearpage
\newpage

\appendix
\section{Simple Confounding} \label{app:simple}

\begin{align}
p(Y(a)) &= \sum_C p(Y(a) \mid C)p(C) \\
&= \sum_C  p(Y(a) \mid A, C)p(C) \label{eq:simple1} \\
&= \sum_C p(Y \mid A, C)p(C) \label{eq:simple2} 
\end{align}

Eq \ref{eq:simple1} holds because $Y(a) \perp A \mid C$, as seen in Figure \ref{fig:simple1}.
Plugging this distribution into $\tau_S = E[Y(1)] - E[Y(0)]$
gives us the causal effect presented in Figure \ref{fig:equations}, Eq \ref{eq:simple}.

This assumes that an intervention on $A$ is well-defined; if we {\it did} conduct a randomized control trial,
we {\it could} assign $A=a$ and break $A$'s dependence on $C$.

In general, this step requires that we condition on all ``back-door'' paths between the treatment and the outcome.
In Figure 1(a), if we did not have data on $C$, we could not block the back-door path between $A$ and $Y$.

Eq \ref{eq:simple2} holds due to consistency. We assume that, given we intervened to set $A=a$, if that individual
would have been assigned $A=a$ in nature, then the distribution over $Y$ is the same.

\section{Missing Data} \label{app:missing}

Denote $p(Y(A(1) = a)) = p(Y(a))$.

First, we identify the causal effect in terms of the true $A(1)$.

\begin{align}
p(&Y(a)) \nonumber\\
&= \sum_C p(Y(a) \mid C)p(C) \label{eq:der1a} \\
&= \sum_C p(Y(a) \mid A(1), C)p(C) \label{eq:der1b} \\
&= \sum_C p(Y \mid A(1), C)p(C) \label{eq:der1}
\end{align}

Where \ref{eq:der1a} holds by chain rule, \ref{eq:der1b} holds by
$A(1) \perp Y(a) \mid C$, and \ref{eq:der1} by consistency.

Now, we identify $A(1)$ in terms of observed data.

\begin{align}
p(&A(1), C, Y) \nonumber\\
&= p(A(1) \mid C, Y)p(C, Y) \label{eq:der2a} \\
&= p(A(1) \mid C, Y, R_A = 1)p(C, Y) \label{eq:der2b} \\
&= p(A \mid C, Y, R_A = 1)p(C, Y)  \label{eq:der2}
\end{align}

Where \ref{eq:der2a} holds by chain rule, \ref{eq:der2b} by
$A(1) \perp R_A \mid C, Y$, and \ref{eq:der2} by consistency.

Now, use Eq \ref{eq:der2} to identify $p(Y \mid A(1), C)$ from
Eq \ref{eq:der1} in terms of observed data.

\begin{align}
p(&Y \mid A(1), C) \nonumber\\
&= \dfrac{p(Y, A(1), C)}{p(A(1), C)} \label{eq:der3a} \\
&= \dfrac{p(Y, A(1), C)}{\sum_Y p(Y, A(1), C)} \label{eq:der3b} \\
&= \dfrac{p(A \mid C, Y, R_A = 1)p(C, Y)}{\sum_Y p(A \mid C, Y, R_A = 1)p(C, Y)}
\label{eq:der3c}\\
&= \dfrac{p(A \mid C, Y, R_A = 1)p(Y \mid C)}{\sum_Y p(A \mid C, Y, R_A = 1)p(Y \mid C)}
\label{eq:der3}
\end{align}

Where \ref{eq:der3a} holds by definition, \ref{eq:der3b} holds by marginalization,
\ref{eq:der3c} holds by an application of \ref{eq:der2} twice,
and \ref{eq:der3} holds by canceling out p(C).

If we include text in this derivation, we simply replace $p(A \mid C, Y, R_A = 1)$
with $p(A \mid {\bf T}, C, Y, R_A = 1)$, where ${\bf T}$ is all our text variables.

Finally, combine Eq \ref{eq:der1} and Eq \ref{eq:der3} to get:

{\small
\begin{align}
p(&Y(A(1) = a)) \nonumber\\
&= \sum_C \dfrac{p(A \mid C, Y, R_A = 1)p(Y \mid C)}{\sum_Y p(A \mid C, Y, R_A = 1)p(Y \mid C)}p(C) \label{eq:missing-counterfactual}
\end{align}
}

Plugging this distribution into $\tau_\text{MD} = E[Y(1)] - E[Y(0)]$
gives us the causal effect presented in Figure \ref{fig:equations}, Eq \ref{eq:missing}.

\section{Measurement Error} \label{app:measurement}

Define the following terms for convenience:

{\small
\begin{align}
\epsilon_{c, y} &= p(A=0 \mid A^* = 1, C=c, Y=y)\\
\delta{c, y} &= p(A=1 \mid A^* = 0, C=c, Y=y)\\
q_{c,y}(0)  &= p(C=c, Y=y, A^* = 0)\\
q_{c,y}(1)  &= p(C=c, Y=y, A^* = 1)
\end{align}
}

Eq (5) and (7) from Pearl 2010 gives us:

\begin{align}
p(&A=1, C=c, Y=y) \nonumber \\
&= \dfrac{-\delta_{c, y}q_{c,y}(0) + (1 - \delta_{c,y})q_{c,y}(1)}{(1 - \epsilon_{c, y} - \delta_{c, y})}\\
p(&A=0, C=c, Y=y) \nonumber \\
&= \dfrac{(1-\epsilon_{c, y})q_{c,y}(0) - \epsilon_{c,y} q_{c,y}(1)}{(1 - \epsilon_{c, y} - \delta_{c, y})}
\end{align}

Now, 

\begin{align}
p(&Y \mid A=1, C) \nonumber \\
&= \dfrac{p(Y, A=1, C)}{p(A=1, C)}\\
&= \dfrac{p(Y, A=1, C)}{\sum_Y p(Y, A=1, C)}\\
&= \dfrac{\dfrac{-\delta_{c, y}q_{c,y}(0) + (1 - \delta_{c,y})q_{c,y}(1)}{(1 - \epsilon_{c, y} - \delta_{c, y})}}{\sum_{y'}\dfrac{-\delta_{c, y'}q_{c,y'}(0) + (1 - \delta_{c,y'})q_{c,y'}(1)}{(1 - \epsilon_{c, y'} - \delta_{c, y'})}}
\end{align}

and then,

\begin{align}
p(&Y \mid A=0, C) \nonumber \\
&= \dfrac{p(Y, A=0, C)}{p(A=0, C)}\\
&= \dfrac{p(Y, A=0, C)}{\sum_Y p(Y, A=0, C)}\\
&= \dfrac{\dfrac{(1-\epsilon_{c, y})q_{c,y}(0) - \epsilon_{c,y} q_{c,y}(1)}{(1 - \epsilon_{c, y} - \delta_{c, y})}}{\sum_{y'}\dfrac{(1-\epsilon_{c, y'})q_{c,y'}(0) - \epsilon_{c,y'} q_{c,y'}(1)}{(1 - \epsilon_{c, y'} - \delta_{c, y'})}}
\end{align}

Plugging this distribution into $\tau_\text{ME} = E[Y(1)] - E[Y(0)]$
gives us the causal effect presented in Figure \ref{fig:equations}, Eq \ref{eq:measurement}.

\newpage

\section{Synthetic Data Distribution} \label{app:synthetic}

In the distributions below, Ber($p$) is used as the abbreviation a Bernoulli distribution
with probability $p$.

Below, $s_i$, $u_i$ and $v_i$ are the effect of C, A, and Y on the probability
of word $T_i$; each is drawn from $\mathcal{N}(0, \zeta)$,
a parameter which controls how correlated words are with the underlying variables.
When $\zeta$ is close to 0, the words are essentially random. When $\zeta$ is large,
the words are essentially deterministic functions of the underlying variables.
Similarly $w_i$ is the effect of word $T_i$ on $R_A$, and is drawn from
$\mathcal{N}(0, \eta)$.

For both settings, we set vocabulary size to 4,334 (to match Yelp experiments) and
$\zeta$ = 0.5. For the missing data setting, we set $\eta$ = 0.1.
We picked these constants by empirically finding a reasonable middle ground
between the text data providing only noise and being a deterministic
function of their parents.
We picked all other constants such that the na\"ive correlation $p(Y \mid A)$
was a poor estimate of the counterfactual $p(Y(a))$ in the full-data setting.

\subsection{Missing data data-generation}

\begin{align*}
C &\sim \text{Ber}(0.4)\\
A(1) &\sim \text{Ber}(-0.3C + 0.4)\\
Y &\sim \text{Ber}(0.2C + 0.1A + 0.5)\\
T_i &\sim \text{Ber}(0.5 + u_iA + v_iC)\\
R_A &\sim \text{Ber}\left(0.7 + 0.2C - 0.4Y + \sum_i w_{i}T_i\right)
\end{align*}

\subsection{Measurement error data-generation}

\begin{align*}
C &\sim \text{Ber}(0.4)\\
A &\sim \text{Ber}(-0.3C + 0.4)\\
Y &\sim \text{Ber}(0.2C + 0.1A + 0.5)\\
T_i &\sim \text{Ber}(0.5 + s_iC + u_iA + v_iY)\\
\end{align*}

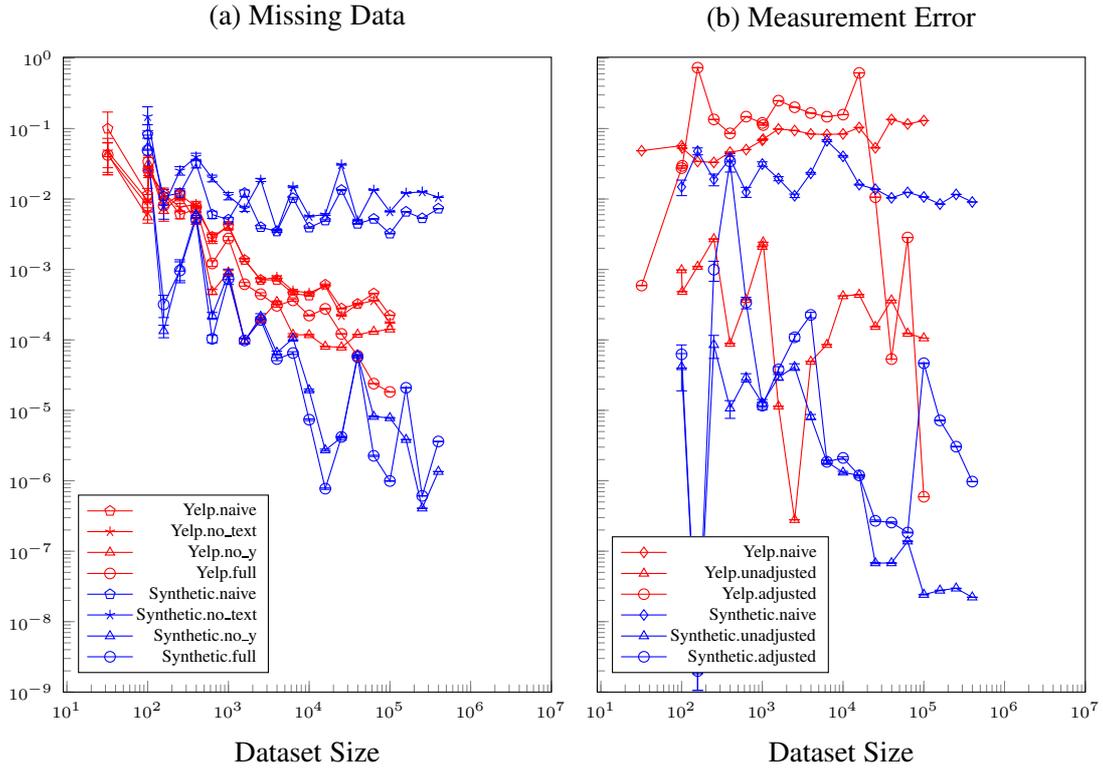
\begin{figure*}[t]
\begin{center}
\centering
\begin{tikzpicture}
\begin{loglogaxis}[
title={(a) Missing Data},
xlabel={Dataset Size},
height=10cm,
width=8cm,
ymin=.000000001,
ymax=1.03,
xmax=1e7,
ytick pos=left,
xtick pos=left,
ticklabel style={font=\tiny},
legend entries={
                Yelp.naive, Yelp.no\_text, Yelp.no\_y, Yelp.full,
                Synthetic.naive, Synthetic.no\_text, Synthetic.no\_y, Synthetic.full,
                },
legend style={ nodes={scale=0.6, transform shape}, cells={anchor=east}, legend pos=south west,},
]
\addplot [color=red, mark=pentagon,]
 plot [error bars/.cd, y dir = both, y explicit]
 table[x =n, y =err, y error =se]{arxiv_dats/md.yelp.10.naive.dat};
\addplot [color=red, mark=star,]
 plot [error bars/.cd, y dir = both, y explicit]
 table[x =n, y =err, y error =se]{arxiv_dats/md.yelp.10.textless.dat};
\addplot [color=red, mark=triangle,]
 plot [error bars/.cd, y dir = both, y explicit]
 table[x =n, y =err, y error =se]{arxiv_dats/md.yelp.10.bad_mi.dat};
\addplot [color=red, mark=o,]
 plot [error bars/.cd, y dir = both, y explicit]
 table[x =n, y =err, y error =se]{arxiv_dats/md.yelp.10.mi.dat};
\addplot [color=blue, mark=pentagon,]
 plot [error bars/.cd, y dir = both, y explicit]
 table[x =n, y =err, y error =se]{arxiv_dats/md.synthetic.10.naive.dat};
\addplot [color=blue, mark=star,]
 plot [error bars/.cd, y dir = both, y explicit]
 table[x =n, y =err, y error =se]{arxiv_dats/md.synthetic.10.textless.dat};
\addplot [color=blue, mark=triangle,]
 plot [error bars/.cd, y dir = both, y explicit]
 table[x =n, y =err, y error =se]{arxiv_dats/md.synthetic.10.bad_mi.dat};
\addplot [color=blue, mark=o,]
 plot [error bars/.cd, y dir = both, y explicit]
 table[x =n, y =err, y error =se]{arxiv_dats/md.synthetic.10.mi.dat};
\end{loglogaxis}
\end{tikzpicture}
\hspace{-0.5em}
\begin{tikzpicture}
\begin{loglogaxis}[
title={(b) Measurement Error},
xlabel={Dataset Size},
height=10cm,
width=8cm,
ymin=.000000001,
ymax=1.03,
xmax=1e7,
ytick pos=left,
yticklabels={,,},
xtick pos=left,
ticklabel style={font=\tiny},
legend entries={Yelp.naive, Yelp.unadjusted, Yelp.adjusted,
                Synthetic.naive, Synthetic.unadjusted, Synthetic.adjusted},
legend style={ nodes={scale=0.6, transform shape}, cells={anchor=east}, legend pos=south west,},
]
\addplot [color=red, mark=diamond,]
 plot [error bars/.cd, y dir = both, y explicit]
 table[x =n, y =err, y error =se]{arxiv_dats/me.yelp.10.naive.dat};
\addplot [color=red, mark=triangle,]
 plot [error bars/.cd, y dir = both, y explicit]
 table[x =n, y =err, y error =se]{arxiv_dats/me.yelp.10.misspecified.dat};
\addplot [color=red, mark=o,]
 plot [error bars/.cd, y dir = both, y explicit]
 table[x =n, y =err, y error =se]{arxiv_dats/me.yelp.10.correct.dat};
\addplot [color=blue, mark=diamond]
 plot [error bars/.cd, y dir = both, y explicit]
 table[x =n, y =err, y error =se]{arxiv_dats/me.synthetic.10.naive.dat};
\addplot [color=blue, mark=triangle,]
 plot [error bars/.cd, y dir = both, y explicit]
 table[x =n, y =err, y error =se]{arxiv_dats/me.synthetic.10.misspecified.dat};
\addplot [color=blue, mark=o,]
 plot [error bars/.cd, y dir = both, y explicit]
 table[x =n, y =err, y error =se]{arxiv_dats/me.synthetic.10.correct.dat};
\end{loglogaxis}
\end{tikzpicture}
\end{center}
\caption{Experimental results with a vocabulary of size 53,197.
Squared distance (y-axis, lower is better) of the estimated causal effect from
$\tau_\text{SC}$ calculated from the full data with no missing data or measurement error.
Error bars (negligible for larger datasets) are 1.96 times standard error across 10 experiments. \label{fig:results2}}
\end{figure*}

\newpage

\section{Additional Experiments} \label{app:experiments}

Figure \ref{fig:results2} shows the results of a second set of experiments,
which are identical to those described in \S \ref{sec:experiments}
except the vocabulary size is now 53,197 instead of 4,334.
For the Yelp data, the larger vocabulary consists of all words which
appear at least ten times in a sample of 1M reviews.
As the larger vocabulary introduced greater memory requirements,
we did not run these experiments with as large of datasets.

The results of these experiments show roughly the same patterns
as those seen in Figure \ref{fig:results}. The
\texttt{adjusted} measurement error models again appear erratic,  
generally performing worse than the \texttt{unadjusted} models
though better than the \texttt{naive} models.
 
The \texttt{full} missing data model appeared to slightly outperform
the \texttt{no\_y} model on Yelp data but only perform as well on
the synthetic data. Both these models appeared better than the
\texttt{naive} and \texttt{no\_text} models on both datasets.

\end{document}